\documentclass[11pt,a4paper]{article}
\usepackage[hyperref]{acl2019}
\usepackage{times}
\usepackage{latexsym}
\newcommand{\seq}[1]{\mathbf{#1}}
\usepackage{url}
\usepackage{multirow}
\usepackage{enumitem}
\usepackage{CJKutf8}
\usepackage{graphicx}
\aclfinalcopy % Uncomment this line for the final submission
\def\aclpaperid{2442} %  Enter the acl Paper ID here

\title{Fine-Grained Sentence Functions for Short-Text Conversation}

\author{
Wei Bi\textsuperscript{1},
Jun Gao\textsuperscript{1,2},
Xiaojiang Liu\textsuperscript{1},
Shuming Shi\textsuperscript{1}\\
\textsuperscript{1}{Tencent AI Lab, Shenzhen, China}\\
\textsuperscript{2}{School of Computer Science and Technology, Soochow University, Suzhou, China}\\
{\{victoriabi, jamgao, kieranliu, shumingshi\}@tencent.com}
}
\date{}

\begin{document}
\begin{CJK}{UTF8}{gbsn}
\maketitle

\begin{abstract}
 Sentence function is an important linguistic feature referring to a user's purpose in uttering a specific sentence.
 The use of sentence function has shown promising results to improve the performance of conversation models. However, there is no large conversation dataset annotated with sentence functions. In this work, we collect a new Short-Text Conversation dataset with manually annotated SEntence FUNctions (STC-Sefun). Classification models are trained on this dataset to (i) recognize the sentence function of new data in a large corpus of short-text conversations; (ii) estimate a proper sentence function of the response given a test query. We later train conversation models conditioned on the sentence functions, including information retrieval-based and neural generative models. Experimental results demonstrate that the use of sentence functions can help improve the quality of the returned responses.
\end{abstract}

\section{Introduction}

The ability to model and detect the purpose of a user is essential when we build a dialogue system or chatbot that can have coherent conversations with humans.
Existing research has analyzed various factors indicating the conversational purpose such as emotions~\cite{prendinger2005empathic,zhou2018emotional,shi2018sentiment}, topics~\cite{xing2017topic,wang2017steering}, dialogue acts~\cite{liscombe2005using,higashinaka2014towards,zhao2017learning} and so on. 
This work describes an effort to understand  conversations, especially short-text conversations~\cite{shang2015neural}, in terms of {\it sentence function}.
Sentence function is an important linguistic feature referring to a user's purpose in uttering a specific sentence~\cite{rozakis2003complete,ke2018generating}. 
There are four major sentence functions: \textit{Interrogative}, \textit{Declarative}, \textit{Imperative} and \textit{Exclamatory}~\cite{rozakis2003complete}.
Sentences with different sentence functions generally have different structures of the entire text including word orders, syntactic patterns and other aspects~\cite{akmajian1984sentence,yule2016study}.

Some work has investigated the use of sentence function in conversation models. 
For example, \citet{li2016ijcai} propose to output interrogative and imperative responses to avoid stalemates. \citet{ke2018generating} incorporate a given sentence function as a controllable variable into the conditional variational autoencoder (CVAE), which can encourage the model to generate a response compatible with the given sentence function. 

Considering the importance of sentence function in conversation modeling, it is surprised to find that no large conversation dataset has been annotated with sentence functions. In \citet{ke2018generating}, they only labeled a small dataset with 2,000 query-response pairs. Sentence function classifiers are trained and tested on this dataset and the best model only achieves an accuracy of 78\%, which is unsatisfactory to serve as an annotation model to automatically assign sentence functions for unlabeled conversation data.

\begin{figure}
    \centering
    \includegraphics[width=\linewidth]{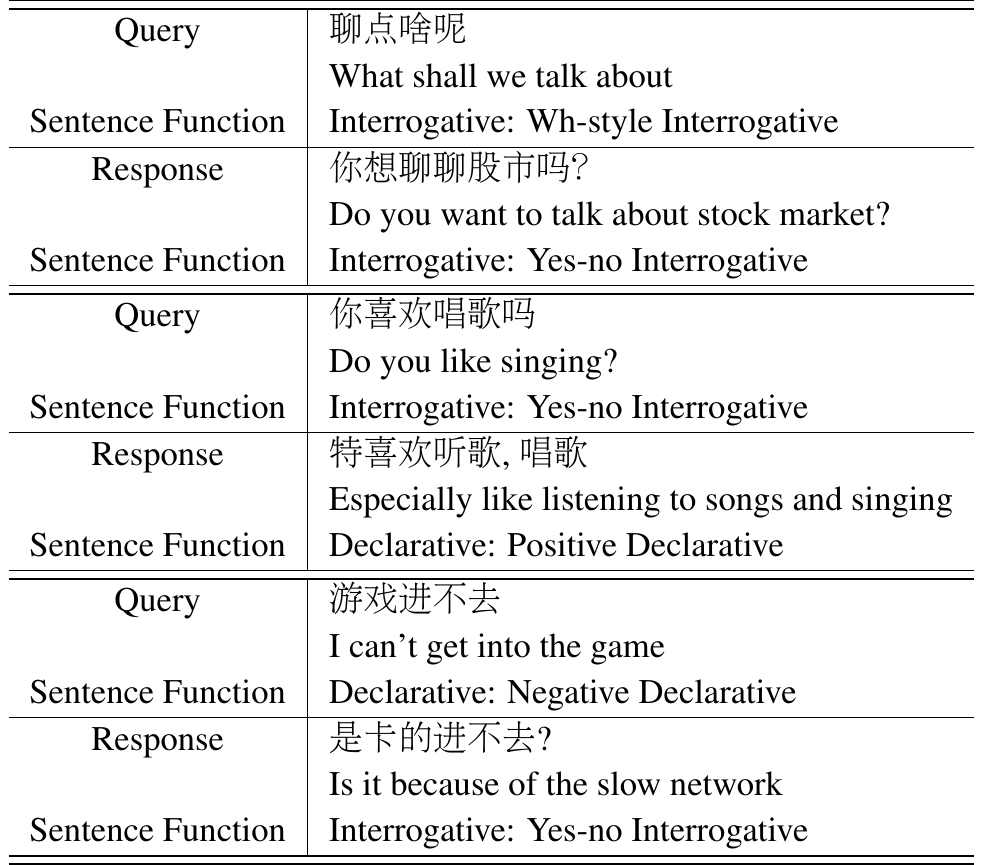}
    \caption{Query-response pairs in the STC-SeFun dataset. The manually annotated level-1 and level-2 sentence functions are separated by the colon.}
    \label{fig:data_example}
\end{figure}

The goal of this work is two fold. On one hand, we create a new Short-Text Conversation dataset with manually annotated SEntence FUNctions (STC-Sefun), in which each sentence segment in the query-response pairs  is labeled with its sentence functions. 
Besides the four major sentence functions, we get inspired by the dialogue act tag set~\cite{stolcke2000dialogue} and further decompose each of them into fine-grained sentence functions according to
their different purposes indicated in conversations. 
For example, \textit{Interrogative} is divided into \textit{Wh-style Interrogative}, \textit{Yes-no Interrogative} and other six types. As shown in the first two examples in Figure~\ref{fig:data_example}, queries expressed in
a \textit{Yes-no Interrogative} sentence
and a \textit{Wh-style Interrogative} sentence
have divergent word patterns and their corresponding responses are also far different.
We have twenty fine-grained sentence functions in total. And we annotate each query-response pair with this two-level sentence function label set. 

%Limitation:
%1 four classes are not enough. fine-grained labels are needed.fan

%2 no large annotated da datasets for chit-chat dialogues

%3 how to output a relevant response with a proper da with a given query. 1 how to decide the da of the response
%2 how to select/generate the response consistent with the predicted da and the given query
% \begin{table}[tb]
%     \centering
%     \scriptsize
%     \begin{tabular}{c|l}
%          %query &  i love the weather today very much, how about you?\\\hline
%          \hline\hline
%          Query & 聊点啥呢 What shall we talk about\\
%          Sentence Function & Interrogative: Wh-style Interrogative\\
%           \hline
%          Response & 你想聊什么 What would you like to talk about\\
%          Sentence Function & Interrogative: Wh-style Interrogative\\\hline\hline
%          Query & 你喜欢唱歌吗 Do you like singing?\\
%          Sentence Function & Interrogative: Yes-no Interrogative\\\hline
%          Response & 特喜欢听歌,	唱歌  Especially like listening to\\
%          &  songs and singing\\
%          Sentence Function & Declarative: Positive Declarative \\\hline\hline
%          Query & 游戏进不去 I can't get into the game\\
%          Sentence Function & Declarative: Negative Declarative\\\hline
%          Response &是卡的进不去? Is it because of the slow network\\
%          Sentence Function & Interrogative: Yes-no Interrogative\\\hline\hline
%     \end{tabular}
%     \caption{Query-response pairs in the STC-SeFun dataset. The manually annotated level-1 and level-2 sentence functions are separated by the colon.}
%     \label{fig:data_example}
% \end{table}

On the other hand, we investigate how to output a response with the consideration of sentence function to improve the performance of conversation models. We decompose this task into two sub-tasks. 
First, we perform two sentence function classification tasks on the STC-SeFun dataset to: (1) determine the sentence functions of unlabeled queries and responses in a large corpus of short-text conversations, and (2) predict a target response sentence function for a given test query. 
Second, we explore various conversation models utilizing sentence function in different manners. These models include information retrieval-based and neural generative models, which are built upon the large automatically annotated corpus and tested with the predicted target sentence function.
We show experimentally that the sentence function classifiers on the two classification tasks achieve sufficiently reliable performance, and sentence function can help improve the relevance and informativeness of the returned responses in different types of conversation models. All our code and datasets are available at \url{https://ai.tencent.com/ailab/nlp/dialogue}.
%Experiments validate the performance on these two sub-tasks.

%%%%%%%%%%%%%%%%%%%%%%%%%%%%%%%%%%%%%%%%%%%
\section{Related Work}
%different from sentiment, topic
Research on dialogue systems or chatbots have studied to control the output responses with different signals to improve user satisfaction of the interaction. 
Various methods consider emotions or topics as the controlling signals.
For example, \citet{martinovski2003breakdown} find that many conversation breakdowns could be avoided if the chatbot can recognize the emotional state of the user and give different responses accordingly.  
\citet{prendinger2005empathic} show that an empathetic responding scheme can contribute to a more positive perception of the interaction.
%
%Besides utilizing emotions as the controlled signal, 
\citet{xing2017topic} observe that users often associate an utterance in the conversation with concepts in certain topics, and a response following a relevant topic could make the user more engaged in continuing the conversation.
The above studies, involving the control of emotions or topics, often affects a few words in the whole returned response, such as \textit{smiling} for the happy emotion, \textit{moisturizing} for the skincare topic. Different from them, sentence function adjusts the global structure of the entire response, including changing word order and word patterns~\cite{ke2018generating}.

\begin{figure*}
    \centering
    \includegraphics[width=\linewidth]{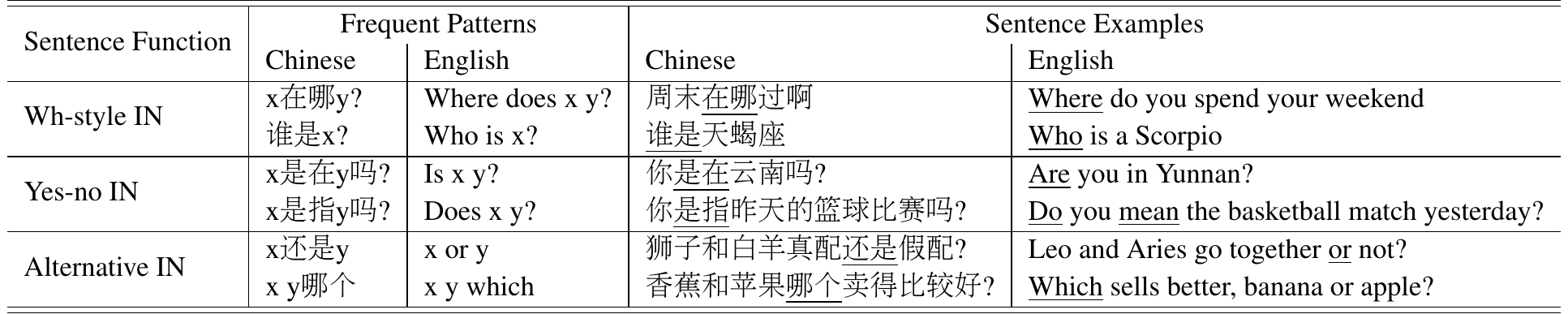}
    \caption{Frequent word patterns of three level-2 \textit{Interrogative} sentence functions. x and y are variables to represent the content words. The underlined words in the sentences are those corresponding to the word patterns.}
    \label{fig:pattern}
\end{figure*}

%different from da:
Modeling dialogue acts such as \textit{statement, question} and \textit{backchannel}, in conversation models has also attracted many researchers' attention. 
~\citet{higashinaka2014towards} identify dialogue acts of utterances, which later contribute to the selection of appropriate responses.
\citet{zhao2017learning} utilize dialogue acts as the knowledge guided attributes in the CVAE for response generation. 
Sentence function is similar to dialogue act in that they both indicate the communicative purpose of a sentence in conversation. 
Moreover, our fine-grained sentence function types are in many ways inspired from the dialogue act tag set~\cite{stolcke2000dialogue} designed for the Switchboard corpus~\cite{godfrey1997switchboard}, which consists of human-human conversational telephone speech. However, the conversations in the Switchboard corpus is multi-round, multi-party and aligned with speech signals. In our work, we target for the single-round non-task-oriented short-text conversation data collected from social media platforms. Thus we remove tags that cannot be determined in our setting, i.e. those needed to be determined in multiple rounds, involved multiple parties, or related to speech signals. Then we merge the remaining tags that have no big difference in their sentence word orders or patterns into one level-2 label. As a result, we have twenty fine-grained sentence functions in our annotation task.

Most existing conversation models can be categorized into two types: the information retrieval~(IR)-based models and the neural generative models.
%Most current conversation models can be categorized into two types:
The IR-based models search for the most similar query in the repository and directly copy its corresponding response as the result~\cite{ji2014information,hu2014convolutional}. 
Meanwhile, the generative models learn the mapping from the input query to the output response in an end-to-end manner~\cite{xing2017topic,zhou2018emotional,zhao2017learning}. 
%
%Early conversation models considering controllable signals are developed as IR-based models~\cite{}. Recent work has shifted on the generative models~\cite{}.
Specifically, \citet{ke2018generating} propose a generative model to deal with the compatibility of controlling sentence function and generating informative content. In our experiments, we use this model as one of the compared methods to analyze the performance on our large conversation corpus.

%existing controllable response selection/generation

%%%%%%%%%%%%%%%%%%%%%%%%%%%%%%%%%%%%%%%%%%%%%%%%%%%%%%
\section{Data Collection}
In this section, we describe the annotation process of the STC-SeFun dataset: (1) how we collect high-quality conversation pairs to be annotated; (2) how we annotate sentence functions for these conversation pairs.

\subsection{Conversation Data Preparation} \label{sec:data_prepare}
We collect a huge number of raw query-response pairs from popular Chinese social media platforms, including Tieba, Zhidao, Douban and Weibo. 
We first pre-process the raw data to filter out pairs that contain dirty words and other sensitive content.
Next, four annotators from a commercial annotation company are recruited to select out high-quality pairs, in which the responses should be not only relevant to the query but also informative or interesting.
Each response is assigned to two different annotators and annotated independently.
We then select out 100k query-response pairs that both annotators consider high-quality for the sentence function annotation task.

\subsection{Sentence Function Annotation}
For a given query-response pair, we first segment the sequence of the query/response by its punctuation. Then we hire three annotators from the same commercial annotation company to annotate sentence functions of each sequence segment.

We design a two-level sentence function label set for annotation. For the level-1 sentence functions, we have the typical four labels: \textit{Declarative~(DE)}, \textit{Interrogative~(IN)}, \textit{Imperative~(IM)} and \textit{Exclamatory~(EX)}.
%, we add two more types that are special sentence functions in social media conversations: \textit{oral-tone and emoticon}， which corresponds to sentences with pure oral-tone words or emoticons that generally express no real meaning but certain attitudes or emotions of the users; and \textit{emoji}, which corresponds to sentences with pure emoji. Sentence segments annotated with these two types should be specifically addressed in many tasks, such as that when we do the word segmentation task to build the vocabulary used in the conversation models. 
%
We further categorize them into the level-2 fine-grained labels due to their different purposes in the conversations.
For example, 
%\textit{declarative} is further divided into \textit{affirmative declarative}, \textit{negative declarative}, \textit{declarative with interrogative words} and other three declarative types. 
\textit{IN} is further divided into \textit{Wh-style IN}, \textit{Yes-no IN} and other six IN types due to the fact that word patterns in different IN labels differ significantly. Figure~\ref{fig:pattern}
illustrates some frequent patterns for these fine-grained IN sentence functions.
%In total, we have twenty-two second-level sentence function types. 
In total, we have twenty level-2 sentence function labels, which are shown in Table~\ref{tab:data_stat}. The explanation of each sentence function is provided in Appendix. 

For each conversation pair, each query/response segment is annotated with both the level-1 and level-2 sentence function labels.
Figure~\ref{fig:data_example} shows three annotated examples. 
The detailed annotation process consists of two stages:
\begin{itemize}[wide=0\parindent]
    \item We ask three annotators to select at most one level-1 label and two level-2 labels for each sentence segment. During annotation, the annotator should consider the query and response jointly to assign the sentence functions.
    \item After all annotators finish labeling the same conversation pair, we re-annotate it as follows: (1) if all three annotators assign the same labels, this data pair is not re-annotated; (2) if labels from all annotators have no overlap or the conversation pair has a sentence segment with no annotated label at all, we ignore this pair; (3) we present all labels together with the majority-voting results back to the annotator who gives the inconsistent label  and ask him/her to check if he/she agrees with the majority-voting results. If this annotator agrees with the majority-voting results, we store this conversation pair with the confirmed results, otherwise we ignore it.
\end{itemize}

\begin{table}[tb]
    \centering
    \small
  %\small
    \setlength{\tabcolsep}{1mm}
    \begin{tabular}{l|l|l}
    \hline\hline
          Sentence Function &  Query & Response \\\hline
          Declarative (DE) & & \\
          \quad Positive DE  & 49,223 (48\%) & 67,540 (57\%)\\
          \quad Negative DE & 9,241(9\%) & 18,428(16\%) \\
          \quad DE with IN words & 887(.9\%) & 2,660(2\%) \\
        % & declarative with heteronyms &  & \\
          \quad Double-negative DE & 40($<$.1\%) & 99(.1\%)\\
          \quad Other types of DE & 2,675(3\%) & 5,218(4\%) \\
          \hline
          Interrogative(IN) & &\\
          \quad Wh-style IN & 23,385(23\%) & 7,652(7\%) \\
          \quad Yes-no IN & 6,469(6\%) & 4,046(3\%) \\
          \quad A-not-A IN & 6,456(6\%) & 1,055(.9\%) \\
          \quad Alternative IN & 789(.8\%) & 279(.2\%) \\
          \quad IN with tag question & 170(.2\%) & 271(.2\%) \\
          \quad Rhetorical & 42($<$.1\%) & 417(.4\%) \\
          \quad IN with backchannel & 0(0\%) & 345(.3\%) \\
          \quad IN with open question & 227(.2\%) & 11($<$.1\%) \\
          %interrogative & interrogative with suggested answers &  & \\
          \hline
          Imperative(IM) & & \\
           \quad IM with request & 2,073(2\%)  & 358(.3\%)\\
          \quad IM with dissuade & 86($<$.1\%) & 58($<$.1\%)\\
          \quad IM with command & 7($<$.1\%) & 4($<$.1\%)\\
          \quad IM with forbidden & 4($<$.1\%) & 2($<$.1\%) \\
          \hline
          Exclamatory(EX) & & \\
          \quad EX without tone words & 241(.2\%) & 3,948(3\%) \\
          \quad EX with interjections & 364(.4\%)  & 1,958(2\%) \\
           \quad EX with greetings & 167(.2\%)  & 285(.2\%) \\
          \hline
       %   Oral-tone(OR)  && \\\hline
       %   Emoji(EM) & &\\\hline
         Total sentences & 95,898 & 95,898\\
         Total sentence segments & 103,138 & 117,714\\\hline\hline
    \end{tabular}
    \caption{Statistics of the SeFun dataset.}
    \label{tab:data_stat}
\end{table}

As a result, we have 95,898 conversation pairs remaining and Table~\ref{tab:data_stat} shows some statistics.

%Different from existing works, we give the DA label for each sentence segment.

%\subsection{Crowdsourcing and Label Aggregation}

%\subsection{Dataset Statistics}
%3 human annotators.
%\begin{table}[htbp]
%    \centering
%    \begin{tabular}{l|c|c}
%         function &  query & response \\\hline
%         declarative & 62,660 & 95,926\\
%         interrogative & 37,715 & 14,405\\
%         imperative & 2,128 & 408\\
%         exclamatory & 456 & 5,133 \\
%         oral-tone & 177 & 1,733\\
%         emoji & 2 & 109\\\hline\hline
%        instance & 95,898 & 95,898\\
%         segment & 102,549 & 114,665\\
%         
%    \end{tabular}
%    \caption{Statistics}
%    \label{tab:my_label}
%\end{table}

%%%%%%%%%%%%%%%%%%%%%%%%%%%%%%%%%%%%%%%%%%%%%%%%%%%%%%%%%%%%%%%%
\section{Sentence Function Classification}
We are given a query with query segments $[\seq{x}_1,\seq{x}_2,\ldots, \seq{x}_n]$, its annotated sentence function labels $[d_{x,1},d_{x,2}, \ldots, d_{x,n}]$, its paired response with response segments $[\seq{y}_1,\seq{y}_2,\ldots, \seq{y}_m]$ and the response sentence function labels $[d_{y,1},d_{y,2}, \ldots, d_{y,m}]$.
However, $n$ and $m$ are 1 for most conversation pairs in our STC-SeFun dataset, which involve short queries and responses generally.
We perform two classification tasks:
\begin{itemize}[wide=0\parindent]
\item Given a query/response sentence segment, we design a model to predict its own sentence function. This model helps us to automatically annotate sentence functions for a large number of unlabeled conversation pairs, which can be used to build conversation models considering with sentence function. We refer this as the {\bf Classification-for-Modeling} task.

\item Given a query and its sentence functions, we aim to predict a proper response sentence function for this query.
This model allows us to select a target sentence function, which will be considered when we decide the output response from the conversation model for a test query. We refer this as the {\bf Classification-for-Testing} task.
\end{itemize}

\subsection{Classification-for-Modeling Task}
\label{sec:class_model}
%We perform a DA classification task to assign the correct DA label $d_{x}$ for a given query segment $x$.
{\bf Training setup:}
For this task, we train and test the models using different data setups:
(1) train with annotated query segments only and test on query segments only; (2) train with annotated response segments only and test on response segments only; (3) mix annotated query and response segments together for training and test on query and response segments respectively.

\noindent
{\bf Network structures:} Our model is a two-level classifier performed in a hierarchical fashion. For the first level, we employ an encoder to obtain a sentence representation $\mathbf{v}_x$ for each input sentence segment, which
%The sentence vector 
%is a high level representation of the input sentence and 
can be used as high-level features for sentence classification. Specifically, the encoder is followed by a fully-connected (FC) layer and a softmax layer to estimate the probability of each level-1 sentence function. Mathematically, the  probability distribution of the level-1 sentence function labels $p(d_{x}^{l1}|\mathbf{x})$ is computed as follows:
%\begin{equation}
\begin{eqnarray}
    \mathbf{v}_x &=& \mbox{Encoder}(\mathbf{x}), \\
    p(d_{x}^{l1}|\mathbf{x}) &=& \mbox{Softmax}(\mbox{FC}(\mathbf{v}_x)). \label{eq:class_level1}
%\end{equation}
\end{eqnarray}
To compute the probability of each level-2 sentence function, we first use an embedding vector $\mathbf{e}_{x}^{l1}$ to represent the level-1 sentence function $d_{x}^{l1}$ estimated by Eq.~\ref{eq:class_level1} from a converged model. 
Then $\mathbf{e}_{x}^{l1}$ is added to the sentence vector $\mathbf{v}_x$ 
%as additional information. 
to compute the probability distribution of the level-2 sentence functions %$p(d_{x}^{l2}|\mathbf{x},d_{x}^{l1})$  can be computed 
as follows:
\begin{equation}
    p(d_{x}^{l2}|\mathbf{x},d_{x}^{l1}) = \mbox{Softmax}(\mbox{FC}(\mathbf{v}_x+\mathbf{e}_{x}^{l1})).
\end{equation}
For encoders, two common implementations are attempted.
The first is a CNN-based encoder commonly used for text classification tasks~\cite{kim2014convolutional}.
The second is a RNN-based encoder which encodes a sequence using a bidirectional GRU. 
% We apply an encoder for the input sequence to obtain the sequence representation, and a sentence function decoder to compute the probabilistic distribution over the possible sentence functions. For the encoder, we try two implementations: the LSTM and CNN. xxxxx \footnote{***fill in the detailed layers}

% For the decoder, we xxxx \footnote{***fill in the detailed layers}
\noindent
{\bf Implementation details:}
The dimension of all hidden vectors is 1024.
All parameters are initialized by sampling from a uniform distribution $[-0.1,0.1]$. The batch size is 128. We use the Adam optimizer with the learning rate 0.0001 and gradient clipping at 5.

\noindent
{\bf Constructing the STC-Auto-SeFun dataset:}
The classification results are shown and analyzed in Section~\ref{sec:expt_classification}. Based on the obtained results, we conclude that the estimated sentence functions by our trained models are highly reliable. Thus we apply the best model to automatically annotating the sentence functions for queries and responses on a large unlabeled data corpus, which contains over 700 million conversation pairs crawled using similar steps in Section~\ref{sec:data_prepare}. This dataset, namely the STC-Auto-SeFun dataset, is later used to build conversation models discussed in Section~\ref{sec:model}.

\subsection{Classification-for-Testing Task}
%For the DA sequence from a query, we learn $p(d_{x}| x)$.
%
%Our second DA classification task is to assign the correct DA label sequence $[d_{y,1},d_{y,2}, \ldots, d_{y,m}]$ for the sentence segments from a response given the query and the query DA labels.
%For the DA sequence from a response, we learn
%$p([d_{y}]| y, x, [d_{x}])$.
%

%We have the second DA prediction task, which is to 
In this task, we aim to predict the response sentence function given a query and its sentence functions, i.e. $p(d_{y}| [\seq{x}], [d_{x}])$.
In this work, we focus on the estimation of a single response sentence function and leave the discussion of multiple response sentence functions in future work.
%Here the DA labels can be either annotated by humans or predicted by the trained DA classification model.

%We first describe the model to learn the mapping from the input query to the first sentence function of the response, i.e. $p(d_{y,1}| [\seq{x}], [d_{x}])$.
We employ an encoder for the input query to obtain the query representation, and an embedding layer followed by a BOW layer to obtain the query sentence function representation. Next, we add these two representations up and feed them into a FC layer followed by a softmax output layer to obtain the probabilistic distribution. The parameter setting is the same as in the Classification-for-Modeling task.

%To obtain a list of response sentence functions, we can extend the above model by learning $p(d_{y,k}| [\seq{x}], [d_{x}], [\seq{y_1},\ldots,\seq{y_{k-1}}], [d_{y,1},\ldots,d_{y,k-1}])$.
%That is to use the query information together with the previous $k-1$ response segments and their corresponding sentence functions to predict the current response sentence function.

%%%%%%%%%%%%%%%%%%%%%%%%%%%%%%%%%%%%%%%%%%%%%%%%%%%%%%%%%%%%%
\section{Conversation Models with Sentence Function} \label{sec:model}
%Most current conversation models can be categorized into two types:
%the retrieval-based models, which search for the most similar query in the repository and directly copy its corresponding response as the result; and the generative models, which directly learn the mapping from the input query to the output response in an end-to-end manner.
In the following, we show how to utilize the sentence functions in both the IR-based models and the generative models.
We use the STC-Auto-SeFun as the retrieval/training corpus in the IR-based/generative models. 
And we focus to predict the response with one target sentence function.

\subsection{IR-based Models}
{\bf IR baseline:} 
We adopt a simple IR model by first finding the most similar query in the retrieval corpus, then utilizing its response as the result. Similarity is measured by the Jaccard index between two bags of words.

\noindent
{\bf Re-ranked model:} We now present a method which demonstrates that sentence function can be used to improve the retrieval-based models. 
%we propose the following algorithm:
%(1)we first 
%predict the first DA label of the first potential response segment for a given query and the query DA labels, i.e. $p(d_{y,0}| [\seq{x}])$ or $p(d_{y,0}| [\seq{x}],[d_{x}] )$;(2) 
We first obtain a set of candidate responses for the IR baseline.
%Next, we predict the score of each candidate response segment being associated with a target sentence function $d_{y,0}$ predicted by the Classification-for-Testing model for the current query $\seq{x}$.
Candidate responses are re-ranked based on whether the candidate is assigned with the target sentence function $d_{y}^*$, which is predicted by the Classification-for-Testing model for the current query $\seq{x}$.
We use the Classification-for-Modeling classifier to predict whether a candidate response is tagged with the target sentence function. If the predicted label is not the target sentence function, this candidate response's score will be penalized with a weight by the Classification-for-Modeling classifier's output probability scaled by a constant.

Specifically, assume the IR baseline $f(x,R) \rightarrow \{s_1,s_2,\ldots,s_{|R|}\}$, where $R$ is the set of candidate responses and the IR baseline outputs a ranked list of scores $\{s_1,s_2,\ldots,s_{|R|}\}$ corresponding to the candidate responses $R=\{r_1,r_2,\ldots,r_{|R|}\}$. 
We then run the Classification-for-Modeling classifier to predict the sentence function $d_{r_i}$ for each candidate response $r_i$ with the probability $p_{r_i}$.
A penalty weight is computed for each candidate as:
\begin{equation}
    s_i^{penalty} = \left\{\begin{array}{ll}
        0 & \mbox{if } y_{r_i} = d_{y}^*, \\
        p_{r_i} & \mbox{otherwise.}  
    \end{array}\right.
\end{equation}
That is, if the candidate response $r_i$ is assigned with the target sentence function, $s_i^{penalty}$ is zero. If it is tagged with any other sentence function,  $s_i^{penalty}$ is the highest probability
of the incorrect sentence function, i.e. $p_{r_i}$.

The re-ranking score is then computed as:
\begin{equation}
    s_i^{re-rank} = s_i - \lambda (s_1 - s_k) s_i^{penalty},
\end{equation}
and the candidate responses are sorted according to their $s_i^{re-rank}$'s.
Here, hyper-parameters $\lambda$ and $k$ are used to control sentence function's influence in re-ranking.
If the top
candidate has a penalty weight of 1.0, then
with $\lambda = 1$, it will be moved to the $k$'th position in
the ranking list. Whereas, $\lambda = 0$ corresponds to no re-ranking. 
%function to rank the candidate response segments by a score function $s_{orig} + p(d_{y,0})$ and select $y_0$ as the one with the highest score; 
%(5) we predict the second DA of the second potential response segment $p(d_{y,1}| [\seq{x}], [d_{x}], d_{y,0})$; (6) we obtain the set of candidate response segments for a given input $[\seq{x}]$ and $y_0$, and repeat the steps so on.

%In Step~1, we first build a classifier from the query $\seq{x}$ to the response DA. xxx
%%%%%hierarchical classification problem.

%In Step~2, we build our retrieval system by xxx

\subsection{Generative Models}
%We aim to train a generative conversation model $p(\seq{y}_0 | \seq{x}, [d_{x}],d_{y_0})$. 
%Considering that the current generative models often generate short and simple responses,  

\noindent
{\bf Seq2seq baseline:} We use a one-layer bi-directional GRU for the encoder, and a one-layer GRU for the decoder 
with soft attention mechanism~\cite{bahdanau2015neural}. Beam search is applied in testing.

\begin{table*}[tbp]
 \small
	%\vspace{0em}
	\begin{center}
			\begin{tabular}{ l | c| c | c | c | c | c}
            \hline\hline
			%	\cmidrule[\heavyrulewidth]{1-4}
				\multirow{2}{*}{Method} & \multicolumn{3}{c|}{level-1 sentence functions} & \multicolumn{3}{c}{level-2 sentence functions} \\
				& Accuracy & Macro-F1 & Micro-F1 & Accuracy & Macro-F1 & Micro-F1 \\\hline
				CNN-encoder (separated) & 97.5 &  87.6 & 97.5 & 86.2 & 52.0 & 86.2 \\\hline
				RNN-encoder (separated) & \textbf{97.6} &  90.9 & \textbf{97.6} & 87.2 & \textbf{65.8} & 87.1 \\\hline
				CNN-encoder (joint) & 97.4 & 87.3 & 97.3 & 86.5 & 51.8 & 86.4 \\\hline
				RNN-encoder (joint) & \textbf{97.6} &  \textbf{91.2} & 97.5 & \textbf{87.6} & 64.2 & \textbf{87.6}\\\hline\hline
			\end{tabular}
    \end{center}
    \caption{Results (\%) on 10,000 test query segments on the Classification-for-Modeling task.}\label{tab:query_model}
\end{table*}

\begin{table*}[tbp]
 \small
	%\vspace{0em}
	\begin{center}
			\begin{tabular}{ l | c| c | c | c | c | c}
            \hline\hline
			%	\cmidrule[\heavyrulewidth]{1-4}
				\multirow{2}{*}{Method} & \multicolumn{3}{c|}{level-1} & \multicolumn{3}{c}{level-2} \\
				& accuracy & macro-F1 & micro-F1 & accuracy & macro-F1 & micro-F1 \\\hline
				CNN-encoder (separated) & 95.2 & 76.6  & 95.1 & 79.0 & 43.3 & 79.0 \\\hline
				RNN-encoder (separated) & 95.5 &  85.0 & 95.5 & 80.0 & \textbf{54.2} & 80.0 \\\hline
				CNN-encoder (joint) & 95.2 & 78.4 & 95.2 & 80.3 & 46.0 & 80.2 \\\hline
				RNN-encoder (joint) & \textbf{95.8} & \textbf{85.9} & \textbf{95.8} & \textbf{80.6} & 53.4 & \textbf{80.6} \\\hline\hline
			\end{tabular}
    \end{center}
    \caption{Results (\%) on 10,000 test response segments on the Classification-for-Modeling task.}\label{tab:response_model}
\end{table*}

\noindent
{\bf C-Seq2seq}~\cite{ficler2017controlling}:
We modify the Seq2seq baseline by adding the sentence function embedding as another input at each decoding position. 
%The dimension of the sentence function embedding is xxxx.

\noindent
{\bf KgCVAE}~\cite{zhao2017learning}: 
The basic CVAE introduces a latent variable $z$ to capture the latent distribution over valid responses and optimizes the variational lower bound of the conditional distribution $p(\seq{y}, z | \seq{x})$.
To further incorporate the knowledge-guided features $l$, the KgCVAE assumes that the generation of $\seq{y}$ depends on $z$, $\seq{x}$ and $l$, and $l$ relies on $\seq{x}$ and $z$. The variational lower bound is then revised to consider $l$ jointly.
%We adjust the original KgCVAE which is for multi-round conversation for our single-round setting and 
Here, we use the response sentence function $d_y$ of each conversation pair as the knowledge-guided features.

\noindent
{\bf SeFun-CVAE}~\cite{ke2018generating}:
This model is specifically designed to deal with the compatibility of the response
sentence function $d_y$ and informative content in
generation. It optimizes the variational lower bound of $p(\seq{y}, z | \seq{x}, d_y)$, where $z$ is a latent variable assumed to be able to capture the sentence function of $\seq{y}$. Thus a discriminator is added to constrain that the encoding information from $z$ can well realize its corresponding sentence function $d_y$.
% The decoder is also revised to generate words among three types: function-related, topic and ordinary words.

\subsection{Implementation Details}
For the re-ranked IR-based model, we collect the top-20 candidates for re-ranking.
We set $\lambda=1$ and $k=20$.
For all generative models, we use a vocabulary of 50,000 words (a mixture of Chinese words and characters), which covers 99.98\% of words in the STC-Auto-SeFun dataset. All other words are replaced with $<$UNK$>$.
The network parameter setting is identical to the classification task.
%The dimension of all hidden vectors is 1024.
%All parameters are initialized by sampling from a uniform distribution $[-0.1,0.1]$. The batch size is 128. We use the Adam optimizer with the learning rate 0.0001 and gradient clipping at 5.
During testing, we use beam search with a beam size of 5.

%%%%%%%%%%%%%%%%%%%%%%%%%%%%%%%%%%%%%%%%%%%%
%\section{Experiments}
%In this section, we will report
\section{Experiments on Sentence Function Classification}

\subsection{Metrics}
We report \textit{Accuracy} (the percentage of samples with corrected sentence functions), \textit{Macro-F1} (the F1 score that weights equally all classes) and \textit{Micro-F1} (the F1 score that weights equally all test samples).

\subsection{Classification-for-Modeling Task} \label{sec:expt_classification}
\noindent
We randomly sample 10,000 query and response segments respectively from the STC-SeFun dataset for testing.
Results on test query and response segments are summarized in Table~\ref{tab:query_model} and~\ref{tab:response_model} respectively. 
As stated in Section~\ref{sec:class_model}, we train different models with query/response data only (denoted as separated), as well as query and response data jointly (denoted as joint) and try two sentence encoders: CNN-based and RNN-based. 
From the results, we can see that:

\begin{table*}[htbp]
 \small
	%\vspace{0em}
	\begin{center}
			\begin{tabular}{ l | c| c | c | c | c | c}
            \hline\hline
			%	\cmidrule[\heavyrulewidth]{1-4}
				\multirow{2}{*}{Method} & \multicolumn{3}{c|}{level-1} & \multicolumn{3}{c}{level-2} \\
				& Accuracy & Macro-F1 & Micro-F1 & Accuracy & Macro-F1 & Micro-F1 \\\hline
				CNN-encoder (without query SeFun) & 81.2 & 15.1 & 81.1 & 55.7 & 23.5 & 55.7 \\\hline
				RNN-encoder (without query SeFun) & 77.9 &  \textbf{30.3} & 77.9 & \textbf{65.6} & 25.8 & 65.5 \\\hline
				CNN-encoder (with query SeFun) & 81.2 & 17.4 & 81.1 & \textbf{65.6} & 21.1 & 65.6\\\hline
				RNN-encoder (with query SeFun) & \textbf{81.3} & 28.5 & \textbf{81.5} & 65.5 & \textbf{25.7} & \textbf{65.7}\\\hline\hline
			\end{tabular}
    \end{center}
    \caption{Results(\%) on 5,000 test queries on the Classification-for-Testing task.} \label{tab:classify_test}
\end{table*}

\begin{itemize}[wide=0\parindent]
\item The RNN-based encoder is better than the CNN-based encoder on both test query and response segments consistently on all metrics; 
\item There is very little performance difference between the separated and joint training data setting under the same network structure;
\item Accuracy of all models, even for level-2 sentence functions, are much higher than 78\% reported in \citet{ke2018generating}, in which the classifier is for 4-class classification and tested on 250 sentences only. It means our models are more reliable to assign sentence function labels to unlabeled conversation pairs; 
\item Macro-F1 scores, especially for level-2 sentence functions, are much lower than Micro-F1 scores. This indicates our models may not perform well on all sentence functions. However, considering that our conversation data are naturally imbalanced and dominated by a few labels, which can be observed from the statistics in Table~\ref{tab:data_stat}, it is sufficient to discriminate between top classes. 
%We select the RNN-encoder(joint) as the best model and compute its macro-F1 using the top-6 frequent level-2 sentence functions ($>3\%$) in the STC-SeFun dataset only and the score increases to  for test queries and 56.0\% for test responses. 
\end{itemize}
\noindent
Based on the above analysis, we consider the RNN-encoder(joint) as the best model for this task, and apply it for the construction of the STC-Auto-SeFun dataset and the conversation models.

\subsection{Classification-for-Testing Task}
We utilize classifiers for this task to estimate the proper response sentence function given the query with/without the query sentence functions. We also implement the RNN-based and CNN-based encoders for the query representation for comparison. Table~\ref{tab:classify_test} shows the results on 5,000 test queries by comparing the predicted response sentence function with its annotated groundtrue response sentence function.
We can observe that:
\begin{itemize}[wide=0\parindent]
    \item Encoding query sentence functions is useful to improve the performance for both CNN-based and RNN-based encoders.
    \item The RNN-based encoder again outperforms the CNN-based encoder, except for very few cases.
    \item Performance on this task decreases significantly compared to the Classification-to-Modeling task. This is because that this task is more subjective and there may be no definite response sentence function to reply to a given query.
\end{itemize}
Note that in previous work about estimating the next dialogue act from a 33 dialogue act tag set given the context with its dialogue acts ~\cite{higashinaka2014towards}, the models achieve about 28\% on Accuracy. Comparing with them, we consider our model has sufficient
ability to choose a reasonable target response sentence function for a given test query. Here, we choose to use the RNN-encoder (with query SeFun) in the testing of the conversation models discussed in the next section.

%%%%%%%%%%%%%%%%%%%%%%%%%%%%%%%%%%
\section{Experiments on Conversation Models}
\subsection{Metrics}
Since automatic metrics for open-domain conversations may not be consistent with human perceptions~\cite{liu2016not}, we hire three annotators from a commercial annotation company to evaluate the top-1 responses on 200 sampled test queries in four aspects:
\textit{Fluency} (whether a response is grammatical), \textit{Relevance} (whether a response is a relevant
reply to its query), \textit{Informativeness} (whether the response provides meaningful information via some specific words relevant to the query) and \textit{Accuracy} (whether the response is coherent with the target sentence function). Each aspect is graded independently in five grades from 0 (totally unacceptable) to 5 (excellent). We further normalize the average scores over all samples into $[0,1]$.

\subsection{IR-based Models}

\noindent
 Results are shown in Table~\ref{tab:result_ir}. We can make the following observations:
\begin{itemize}[wide=0\parindent]
 %   \item Hit@1 of the re-ranked IR model is not 100\%. This is because a few groundtrue responses are not predicted with the target response sentence function, thus suffer the penalty scores and do not rank at the highest position. 
    %Since we use the Jacquard similarity, the groundtrue response is with the highest similarity score, it is not surprised it rank the first in most cases.
    \item The re-ranked models achieve higher accuracy on the target response sentence function than the IR baselines, which means our designed re-ranking score function is effective.  
    \item For both sentence function levels, the re-ranked IR models perform better than the IR baselines on all metrics. This means that considering a proper sentence function into the IR-based models is useful to help select high-quality responses.
\end{itemize}

\begin{table}[htbp]
 \small
	%\vspace{0em}
	\begin{center}
			\begin{tabular}{ l | c| c |c|c}
            \hline\hline
			%	\cmidrule[\heavyr/ulewidth]{1-4}
				Method  & Flue
				& Rele & Info & Accu  \\\hline
				IR baseline (level1)   & 63.4 &68.4 & 61.5 & 34.3\\\hline
				Re-ranked IR  (level1)  & \textbf{69.6} & \textbf{74.4} & \textbf{77.2} & \textbf{50.5}\\\hline\hline
				IR baseline (level2)  & 63.0& 68.2&61.6&25.0 \\\hline
				Re-ranked IR  (level2)  & \textbf{68.0} & \textbf{73.4} & \textbf{75.3}& \textbf{38.6}\\\hline\hline
			\end{tabular}
    \end{center}
    \caption{Results(\%) on the IR-based models.} \label{tab:result_ir} 
\end{table}

\begin{table}[htbp]
 \small
	%\vspace{0em}
	\begin{center}
			\begin{tabular}{ l |c|c|c|c}
            \hline\hline
			%	\cmidrule[\heavyrulewidth]{1-4}
				Method  &  Flue & Rele & Info & Accu  \\\hline
				%IR baseline & 0.58 & 0.35&0.24 &0.17 &0.287 &0.753 &&\\\hline
				%IR DA-reranked  & 0.23 &0.04 &  0.01& 0.004& 0.265& 0.726& &\\\hline\hline
				Seq2seq(level1)  &55.4 &61.5 &49.3&32.0\\\hline
				C-Seq2seq(level1)   & 55.9 & \textbf{65.0} & \textbf{51.6} &33.0 \\\hline
			    KgCVAE(level1)   &\textbf{57.6}&62.5& 51.4&29.0\\\hline
				SeFun-CVAE(level1)   &57.1 &63.5 &50.9&\textbf{34.5}\\\hline\hline
				Seq2seq(level2) &53.0 & 62.3&48.9& 35.0\\\hline
				C-Seq2seq(level2)  & \textbf{58.9}& \textbf{64.7}& \textbf{50.9} &\textbf{37.2} \\\hline
			    KgCVAE(level2)   &56.5 & 63.2&49.4&33.7 \\\hline
				SeFun-CVAE(level2)   & 56.9 & 63.7 & 50.2 & 36.7\\\hline\hline
			\end{tabular}
    \end{center}
    \caption{Results(\%) of the generative models.}
    \label{tab:result_gen}
\end{table}

\subsection{Generative Models}
\noindent
Results are shown in Table~\ref{tab:result_gen}. We have the following observations:
\begin{itemize}[wide=0\parindent]
    \item For level1 sentence functions, the C-Seq2seq achieves the highest scores on relevance and informativeness, the second best score on accuracy and the third score on fluency. For level2 sentence functions, it performs the best on all metrics. Thus, we consider the C-Seq2seq has the best performance on our test set overall. 
    \item The Seq2seq baseline is inferior to C-Seq2seq and SeFun-CVAE on all metrics. This indicates that with proper use of sentence function, the generative conversation models can effectively improve their performance. 
    \item The KgCVAE obtains the lowest accuracy and we conjecture that the KgCVAE can not effectively capture the sentence function information. By contrast, the accuracy obtained by the SeFun-CVAE, in which a type discriminator is added, is higher than that of the KgCVAE. This shows that the added discriminator in the SeFun-CVAE can effectively constrain the encoding information from the latent variable to well represent its corresponding sentence function.
    \item Compared with two IR-based models, all generative models obtain lower scores on fluency, relevance and informativeness.
    %This is consistent with human perceptions since IR-based models output real responses collected from users from a .
    %Relevance scores of generative models are higher than the IR-baseline but worse than the re-ranked IR model.
    On accuracy, the best generative models, i.e. SeFun-CVAE(level1) and C-Seq2seq(level2) outperform the IR baselines respectively, but still underperform the re-rank IR models.  
    Thus considering all metrics together, the re-ranked IR model performs the best.
\end{itemize}

\begin{figure}[t]
    \centering
    \includegraphics[width=\linewidth]{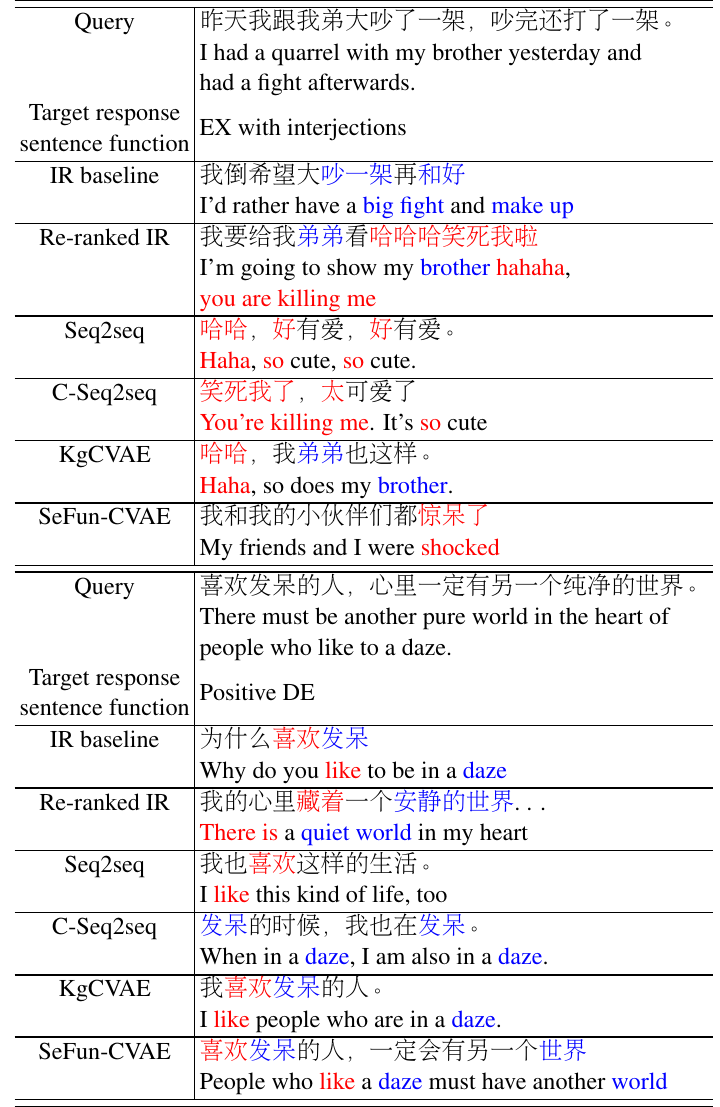}
    \caption{Responses of IR-based and generative models. Words in red are related to the target sentence function and words in blue are relevant to the query.}
    \label{fig:gen_example}
\end{figure}

\subsection{Case Study}
In Figure~\ref{fig:gen_example}, we present two examples, each of which shows a  test query with its target level-2 response sentence function (predicted by the Classification-for-Testing model), and the top-1 responses selected by the IR-based models.

We can see that the IR baseline tends to output responses with more overlapped terms with the query due to the use of Jaccard similarity. However, the obtained responses may not be relevant to the query, as shown in the first case.
%the use of sentence functions helps achieve better responses. 
Whereas, the re-ranked IR model can balance between the compatibility  of the target response sentence function and the Jaccard similarity.
Thus its selected responses may not have many term overlapped with the query, but the conversations continue more smoothly and coherently. 
%Figure~\ref{fig:gen_example} shows the generated responses of all compared generative models on two example queries. 

Responses of the Seq2seq baseline are generic and universal that can be used to reply to a large variety of queries. 
The three improved methods tend to generate responses with some words related to the target sentence functions and relevant to the query. 10y
Thus generative models with the use of sentence function can help improve the response quality, though not as significantly as in the IR-based models.

%%%%%%%%%%%%%%%%%%%%%%%%%%%%%%%%%%%%%%%%%%%%%%%%%%%%%%%%%%%
\section{Conclusions}
This work introduces the STC-SeFun dataset, which consists of short-text conversation pairs with their sentence functions manually annotated. 
%For the task of sentence function classification, 
We first show that
classifiers trained on the STC-SeFun dataset can be used to automatically annotate a large conversation corpus with highly reliable sentence functions, as well as to estimate the proper response sentence function for a test query.
Using the large automatically annotated conversation corpus, we train and evaluate both IR-based and generative conversation models, including baselines and improved variants considering the modeling of sentence function in different ways. Experimental results show that the use of sentence function can help improve both types of conversation models in terms of response relevance and informativeness.

%%%%%%%%%%%%%%%%%%%%%%%%%%%%%%%%%%%%%%%%%%%
%\newpage

% include your own bib file like this:  0
%\bibliographystyle{acl}
%\bibliography{acl2018}
\bibliography{ref}
\bibliographystyle{acl_natbib}
\end{CJK}
\end{document}

\begin{CJK}{UTF8}{gbsn}
\maketitle
\section{Definitions of Sentence Functions}
%Labeled in order of rank, you must mark first-level labeling (only one), then second-level labeling (more than one).
%\section{Specific Rules}
\subsection{First-level labels}
\noindent
\textbf{Declarative}: to explain a fact and opinion, including negation and affirmation. 

Feature: flat town, end with a period, with modal particle sometimes.

E.g.: He knows.  He maybe knows. He doesn’t know yet.

\noindent
\textbf{Interrogative}: to ask interrogatives or express doubts. 

Feature: intonation rising, end with interrogative mark.

E.g.: Did he know it?

\noindent
\textbf{Imperative}: to request the other party. 

Feature: subjects are restricted to second person pronouns, first person plurals, and appellations, with verbs.

E.g.: Let’s go! You must go!

\noindent
\textbf{Exclamatory}: to express strong feelings, containing joy, anger, surprise, sadness, etc. 

Feature: sentences with emotional adjectives, end with exclamatory mark.

E.g.:  What a beautiful day! What a sad movie!

\section{Second-level labels}
\subsection{Declarative}
\noindent
\textbf{Positive declarative}: generally unmarked, modal particle can be emphasized or not. 

%Modal particle: 的、了、嘛、啊、呢、罢了

E.g.: Mr. Wang will go to Beijing tomorrow. 
It’s Mr. Wang who goes to Beijing tomorrow. 
His voice like a serenade.
He admitted to a prestigious university unexpectedly.

\noindent
\textbf{Negative declarative}: usually use negative words, like don’t, haven’t.

E.g.: 
He doesn’t eat.
He didn’t eat.
He can’t understand Chinese.
He didn’t understand Chinese.

\noindent
\textbf{
declarative with interrogative words}: to express subjective.

E.g. 
No one knows what is going on.
He can do any tough thing.
We don’t go anywhere today.
I know why he didn't come.

%Heteronyms: in some idioms, the affirmative form of a sentence is the same as the meaning of a negative form.
%E.g. 
%So happy
%Almost lost balance

\noindent
\textbf{Double negation}: double negation is used to express affirmation. Some to strengthen the affirmative, the other to weaken, like noting but, anything but.

E.g.:  
You have no choice but to come out.
You have no choice but to say.
I don't not like that.

\noindent
\textbf{
Others}: except for the above all, including a single word or phrase.

E.g.:  
Tomato and egg
Honour of Kings

\subsection{Interrogative}

\noindent
\textbf{Yes-no interrogative}: the structure is similar to the declarative, the difference in tone. Generally, after the interrogative word is removed, it is still a complete declarative. The answer is yes or no.

E.g.: 
A: Did he know it?
B: Yes, he did.

\noindent
\textbf{Wh-interrogative}: with interrogative pronouns, who, what, how, etc. The answer is concrete and complex.

E.g.:  
A: Who told him?
B: Xiao Ming
A: When are you going?
B: Tomorrow

\noindent
\textbf{Alternative interrogative}: more than two parallel structure choices, "... or..."

E.g.: 
Do you like apples, pears or bananas?

\noindent
\textbf{A-not-A interrogative}: asking two aspects, the other party will choose one, “wiling or not”” can you”

E.g.: 
Would you like to go to the streets?

%interrogative with suggested answers: self-interrogativeing and self-answer, commonly used in interrogative structure.
%E.g. 
%Do you know how old I am? 25!
\noindent
\textbf{Rhetorical}: negative meaning with a positive form, and vice versa.

E.g.: 
This interrogative is do like this, isn’t right?

\noindent
\textbf{Interrogative with backchannel}: repeat the other party's interrogative, ask for confirmation, to win time to consider how to answer.

E.g.: 
A: What is your last name?
B: My last name?

\noindent
\textbf{Interrogative with tag question}: attached to other sentences (usually declarative sentences), like “right?”” OK?”

E.g.:  
You promise me (affirmative declarative), ok? (tag interrogative)
I do this, right?

\noindent
\textbf{Interrogative with open question}: leading a topic and discussion, can't answer with "yes", "no" or a certain word or number.

E.g.: 
Tell me about your dream!

\subsection{Imperative}
\noindent
\textbf{Command}: strong tone, short declarative, no modal particle

E.g.:  
Hurry up!

\noindent
\textbf{Request}: emotional euphemism, more soothing, use "please" or modal particle.

E.g.: 
Please care for the environment.

\noindent
\textbf{Forbidden}: strong tone, no mood words, use " no”.

E.g.: 
No littering!

\noindent
\textbf{Dissuade}: emotional euphemism, more soothing, use "please not".

E.g.: 
Please not climbing the railings!

\subsection{Exclamatory}
\noindent
\textbf{Exclamatory sentences with interjections}

E.g.: 
Hey! help!
How good is that!
Get out!
Oh my god!

\noindent
\textbf{Exclamatory sentences with greetings}

E.g.: 
Long live the great unity of all ethnic groups!
Come on!
Congratulations!

\noindent
\textbf{Exclamatory sentences without tone particles}

E.g.: 
A: I want to sing again. 
B: me too!
A: I'm going to there.
B: OK!

\end{CJK}